\pdfoutput=1

\documentclass[letterpaper, 10 pt, conference]{ieeeconf}  

\IEEEoverridecommandlockouts                              
\usepackage{epsfig} 
\usepackage{amsmath} 
\usepackage{algorithm2e}
\usepackage{framed}
\usepackage{graphicx}
\usepackage{xcolor}
\usepackage{tikz}

\definecolor{shadecolor}{rgb}{1, 0.8, 0.8}
\definecolor{red}{rgb}{0,0,0}

\title{\LARGE \bf
Automatic Coverage Selection for Surface-Based Visual Localization
}

\author{James Mount$^{1}$, Les Dawes$^{1}$ and Michael Milford$^{1}$
\thanks{$^{1}$J. Mount, L. Dawes and M. Milford are with Faculty of Science and Engineering,
        Queensland University of Technology, Brisbane, Australia
        {\tt\small j.mount@qut.edu.au, l.dawes@qut.edu.au, michael.milford@qut.edu.au}}%
}

\newcommand\copyrighttext{%
  \footnotesize \textcopyright © 2019 IEEE. Personal use of this material is permitted. Permission from IEEE must be obtained for all other uses, in any current or future media, including reprinting/republishing this material for advertising or promotional purposes, creating new collective works, for resale or redistribution to servers or lists, or reuse of any copyrighted component of this work in other works. 
  }
\newcommand\copyrightnotice{%
\begin{tikzpicture}[remember picture,overlay]
\node[anchor=south,yshift=15pt] at (current page.south) {\fbox{\parbox{\dimexpr\textwidth-\fboxsep-\fboxrule\relax}{\copyrighttext}}};
\end{tikzpicture}%
}

\begin{document}

\RestyleAlgo{boxruled}

\maketitle

\copyrightnotice

\thispagestyle{empty}
\pagestyle{empty}

\begin{abstract}

Localization is a critical capability for robots, drones and autonomous vehicles operating in a wide range of environments. One of the critical considerations for designing, training or calibrating visual localization systems is the coverage of the visual sensors equipped on the platforms. In an aerial context for example, the altitude of the platform and camera field of view plays a critical role in how much of the environment a downward facing camera can perceive at any one time. Furthermore, in other applications, such as on roads or in indoor environments, additional factors such as camera resolution and sensor placement altitude can also affect this coverage. The sensor coverage and the subsequent processing of its data also has significant computational implications. In this paper we present for the first time a set of methods for automatically determining the trade-off between coverage and visual localization performance, enabling the identification of the minimum visual sensor coverage required to obtain optimal localization performance with minimal compute. We develop a localization performance indicator based on the overlapping coefficient, and demonstrate its predictive power for localization performance with a certain sensor coverage. We evaluate our method on several challenging real-world datasets from aerial and ground-based domains, and demonstrate that our method is able to automatically optimize for coverage using a small amount of calibration data. We hope these results will assist in the design of localization systems for future autonomous robot, vehicle and flying systems.

\end{abstract}

\section{INTRODUCTION}
Over the past two decades, robotics and autonomous vehicle systems have \textcolor{red}{increasingly utilized} vision sensors, using them to provide critical capabilities including localization. This usage is due in part to the rapid increase in both \textcolor{red}{camera capabilities and computational processing power}. Cameras have benefits over other sensors such as radar, providing far more information about the environment including texture and colour. Furthermore, cameras have other advantages including being passive sensing modalities, \textcolor{red}{and the potential} to be relatively inexpensive, have  small form factors and relatively low power consumption \cite{milford2014condition}.

One of the critical system design considerations for camera-equipped autonomous platforms is the coverage of the cameras, which is affected by a range of factors including the altitude of the platform (for aerial contexts), mounting point (for ground-based vehicles), the camera field of view and the sensor resolution. The choices made with regards to these system properties can also affect other critical system considerations like compute -- if a subset of the entire field of view of a camera can be used for effective localization, significant reductions in compute can be achieved.

We addresses this challenge by presenting a novel technique that automatically identifies the trade-off between visual sensor coverage and the performance of a visual localization algorithm. The technique enables automatic selection of the minimum visual sensor coverage required to obtain optimal performance -- specifically, optimal localization recall without expending unnecessary compute on processing a larger sensor coverage field than required. We focus our research within the area of vision based surface localization, such as that demonstrated by Kelly et al \cite{Kelly:2000,Kelly:2007} for warehouse localization, Conte and Doherty \cite{conte2009vision} in aerial environments and Hover et al \cite{Hover:2012} in ship hull inspection. We evaluate the proposed method \textcolor{red}{using two surface-based visual localization techniques}, on several challenging real-world aerial and ground-based surface datasets, showing that the technique can automatically select the optimal coverage by using calibration data from environments analogous to the deployment environment.

The paper proceeds as follows. Section \ref{section:RelatedWork} summarizes related works, such as surface-based visual localization and procedures for parameter tuning. Sections \ref{section:Approach} and \ref{section:ExperimentalSetup} provide an overview of the calibration procedure and the experimental setup respectively. The performance of our algorithm and a discussion is presented in Sections \ref{section:Results} and \ref{section:Discussion} respectively.

\begin{figure}[!t]
  \centering
  \includegraphics[scale=0.75]{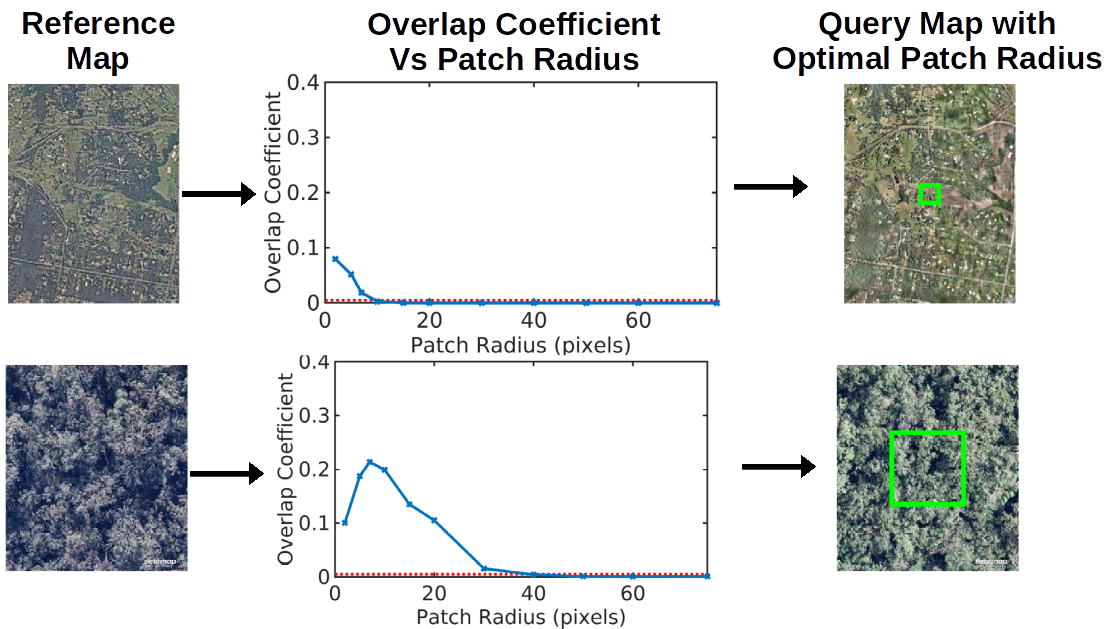}
  \vspace{-0.25cm}
  \caption{Given a reference map and a number of query samples, our overlap coefficient-based calibration process automatically determines the optimal sensor coverage for maximizing localization performance while minimizing computational overhead. The blue and red lines in the plots are the overlapping coefficient for various patch radii for the two datasets shown and the overlapping coefficient threshold respectively.}
  \label{figure:OverviewFig}
\vspace{-0.2cm}
\end{figure}

\section{RELATED WORK} \label{section:RelatedWork}

In this section we present research related to surface-based visual localization and calibration procedures for parameter tuning. \textcolor{red}{The coverage here is of localization techniques themselves rather than coverage calibration approaches; as to the best of our knowledge we do not believe there is a system that is directly comparable to the technique outlined in this paper.}

\subsection{Surface-Based Visual Localization}

In several mobile robotics applications the system moves relative to a surface, such as a drone across the ground, an autonomous vehicle over the road or a submarine relative to a ship's hull. As a result, several approaches have proposed using the surface that the robot moves relative to as a visual reference map for localization. For example, Kelly et al. thoroughly demonstrated that surface-based visual localization using pixel-based techniques for mobile ground platforms is feasible within warehouse environments with controlled lighting using a monocular camera \cite{Kelly:2000,Kelly:2007}. Mount et al. also demonstrated this technique can be applied to autonomous vehicles and a road surface, even with day to night image data \cite{mount2017image}. \textcolor{red}{Additionally, \cite{kozak2016,zhang2019} demonstrate the use of local features for road surface-based visual localization.}

Unmanned aerial vehicles (UAVs) regularly use geo-referenced aerial imagery to help alleviate errors caused by GPS outages \cite{conte2009vision, Sim2002IntegratedPE, caballero2009vision, madison2007vision}. For example, Conte et al. demonstrated that they could incorporate feature-based image registration to develop a drift-free state estimation technique for UAVs \cite{conte2009vision}.

The research presented on underwater visual ship hull inspection and navigation further demonstrates that vision based surface localization is feasible even in challenging conditions \cite{Hover:2012,Kim:2013,Ozog:2014}. There has also been a variety of research into utilizing the surface as the input image stream for visual odometry \cite{Dille:2010,Nourani:2011,aqel2016adaptive}.

\textcolor{red}{All these systems either have a hard-coded empirically tuned parameter defining the amount of the visual sensor to use, or simply use the entire field of view. Therefore, they could be performing unnecessary computations without any performance gains. In contrast, our system automatically selects the optimal visual sensor coverage for maximizing performance while minimizing unnecessary computation.}

\subsection{Calibration Procedures for Visual Localization}

The altering of configuration parameters in both deep learning and traditional computer vision algorithms can have a drastic effect on performance \cite{bergstra2011algorithms}, such as the the size of images used within appearance-based techniques \cite{milford2012seqslam}. This can cause difficulties in successfully making the transition between research and application, as well as between domains \cite{jacobson2018semi, zeng2018i2, zeng2017enhancing}. Due to these difficulties, there have been several research areas investigating the development of automatic calibration routines to improve the performance of visual localization alogrithms. Lowry et al. demonstrated online training-free procedures that could determine the probabilistic model for evaluating whether a query image came from the same location as a reference image, even under significant appearance variation \cite{lowry2014towards, lowry2015building}. In \cite{jacobson2015online, jacobson2015autonomous} and \cite{jacobson2013autonomous} Jacobson et al. explored novel calibration methods to automatically optimize sensor threshold parameters for place recognition. Several bodies of work have also used the system's state estimate to reduce the search space in subsequent iterations, such as that in \cite{aqel2016adaptive, Nourani:2011}. In all bodies of work the authors demonstrated that parameter calibration outperformed their state-of-the-art counterparts. \textcolor{red}{However, these techniques typically focused on optimizing a single metric, mainly recall/accuracy, and did not explicitly consider calibrating for both localization performance and computation load in parallel, which is the focus of the research described in this paper}. 

There has been considerable research into calibration routines to identify spatial and temporal transforms between pre-determined sensor configurations \cite{maddern2012lost, furgale2013unified, kelly2011visual, scaramuzza2007extrinsic, pandey2015automatic, weiss2012real}. Significant investigations into using visual sensors to overcome kinematic and control model errors used in robotics platforms has also been an area of key research \cite{meng2007autonomous, vsvaco2014calibration, du2013online}. These approaches in general have addressed a different set of challenges to those addressed here, instead focusing on the relationship between sensors and robotic platforms or between sensors and other non-localization-based competencies. The automatic selection of hyper-parameters is also related, especially in the deep learning field \cite{bergstra2011algorithms, bergstra2012random, thornton2013auto, bardenet2013collaborative, gold2005bayesian}.

\section{Approach} \label{section:Approach}

This section provides an overview of the approach for automatic selection of the sensor coverage required for an optimal combination of visual surface based localization performance and computational requirements. The primary aim and scope of the techniques presented here is to identify the amount of coverage with respect to the sensor field of view and the altitude of a downward-facing camera above the ground plane. The technique requires a small number of aligned training image pairs from an environment analogous to the deployment environment; although we do not attack that particular problem here, there are a multitude of techniques that could potentially be used to bootstrap this data online such as SeqSLAM \cite{milford2012seqslam}. We outline the complete calibration procedure in Algorithm \ref{algo:CalibrationProcedure}.

\begin{algorithm}
\color{red}
\For{all patch radii in $P_N$}{
    \For{$x$ calibration samples}{
        run localization on sample; \\
        store ground truth and all other localization scores; \\
    }
    fit distribution to ground truth scores; \\
    fit distribution to all other scores; \\
    calculate OVL between distributions; \\
    store patch radius and OVL in matrix;
}
\eIf{any OVL value $\leq$ required OVL value}{
    interpolate to find optimal patch radius; \\
}{
    set optimal patch radius to ${arg\, max}_N{(P_N)};$ \\
}
\caption{Calibration Procedure}
\label{algo:CalibrationProcedure}
\end{algorithm}

\subsection{Optimal Coverage Calibration Procedure}
\begin{figure}[!t]
  \centering
  \includegraphics[scale=0.2]{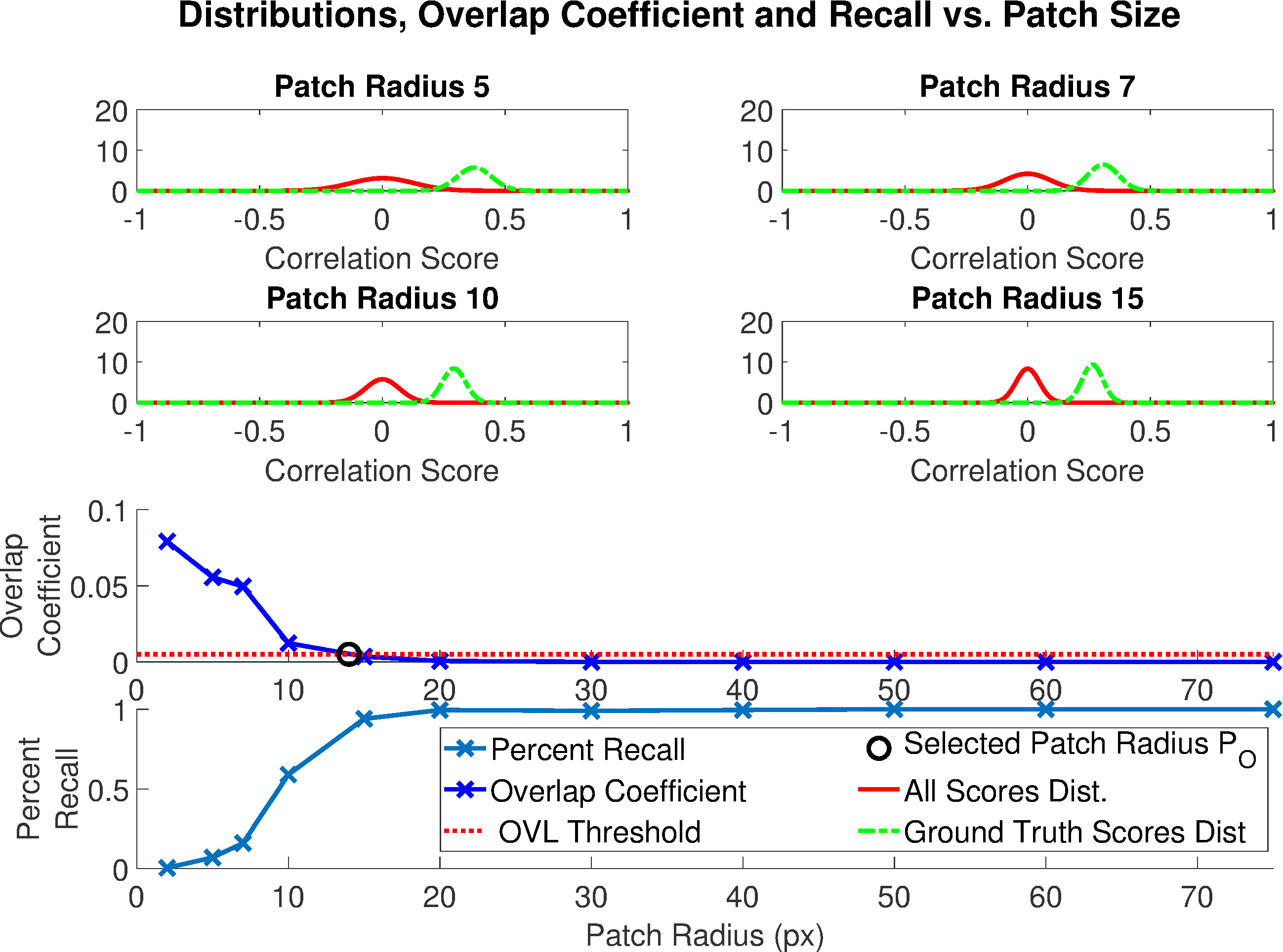}
  \vspace{-0.2cm}
  \caption{The effect of patch radius on the overlapping coefficient (OVL) between the normal distributions of all the correlation scores (solid red line) and the ground truth only scores (dashed green line). The red dotted line and solid black circle in the bottom plot represents the required OVL value $O_r$ and the selected interpolated patch radius respectively. This examples used NCC as the underlying localization technique.}
  \label{figure:OVLMetricExample}
\vspace{-0.2cm}
\end{figure}

The calibration procedure works under the assumption that the similarity of the normal distributions between the ground truth only scores and all scores diverges as sensor coverage, resolution and placement changes. This divergence in distribution similarity is indicative of better single frame matching performance (see Figure \ref{figure:OVLMetricExample} for an example). In this paper we use the Overlapping Coefficient (OVL), which is an appropriate measure of distribution similarity \cite{inman1989overlapping, reiser1999confidence}. There are various measures for OVL, including Morisita's \cite{morisita1959measuring}, Matusita's \cite{matusita1955decision} and Weitzman's \cite{weitzman1970measures}. We use Weitzman's measure which is given by,

\begin{equation}
    O = \int_{k_0}^{k_1} min(p(x), q(x)) dx
\end{equation}

where $p(x)$ and $q(x)$ are two normal distributions and $O$ is the resulting OVL value. \textcolor{red}{The bounds of the integral, $k_0$ and $k_1$, are the numerical limits of the technique being utilised. For example, $k_0$ and $k_1$ would be $-1$ and $1$ respectively for NCC. The Overlapping Coefficient was used as the measure of distribution similarity over other methods, such as the Kullback-Leibler divergence, as it decays to zero as two distributions become more dissimilar and because it is symmetric.}

Once the OVL value goes below a given threshold there is limited to no performance gains in localization performance. It is at this point we consider the visual sensor coverage to be optimal. As the OVL threshold is most likely between two of the tested calibration OVL values, as in Figure \ref{figure:OVLMetricExample}, we use linear interpolation to select the point of intersection. If no tested calibration points achieve less than the required OVL we simply take the largest coverage tested. The selection of the optimal operating value $P_O$ hence is given by the following,

\begin{equation}
\color{red}
    P_O = \begin{cases}
    P_a+(P_b-P_a)\frac{O_r-O_a}{O_b-O_a} & \text{any}(P_N \leq O_r) \\
    {{arg\, max}_N}{(P_N)} & \text{otherwise}
    \end{cases}
\end{equation}

where $P_O$, $P_a$ and $P_b$ are the optimal operating value, and the value above and below the required OVL threshold, $O_r$, respectively. $O_a$ and $O_b$ are the corresponding OVL values for the tested calibration values $P_a$ and $P_b$. $P_N$ are all the values tested during calibration.

\begin{figure}[!t]
  \centering
  \includegraphics[scale=0.085]{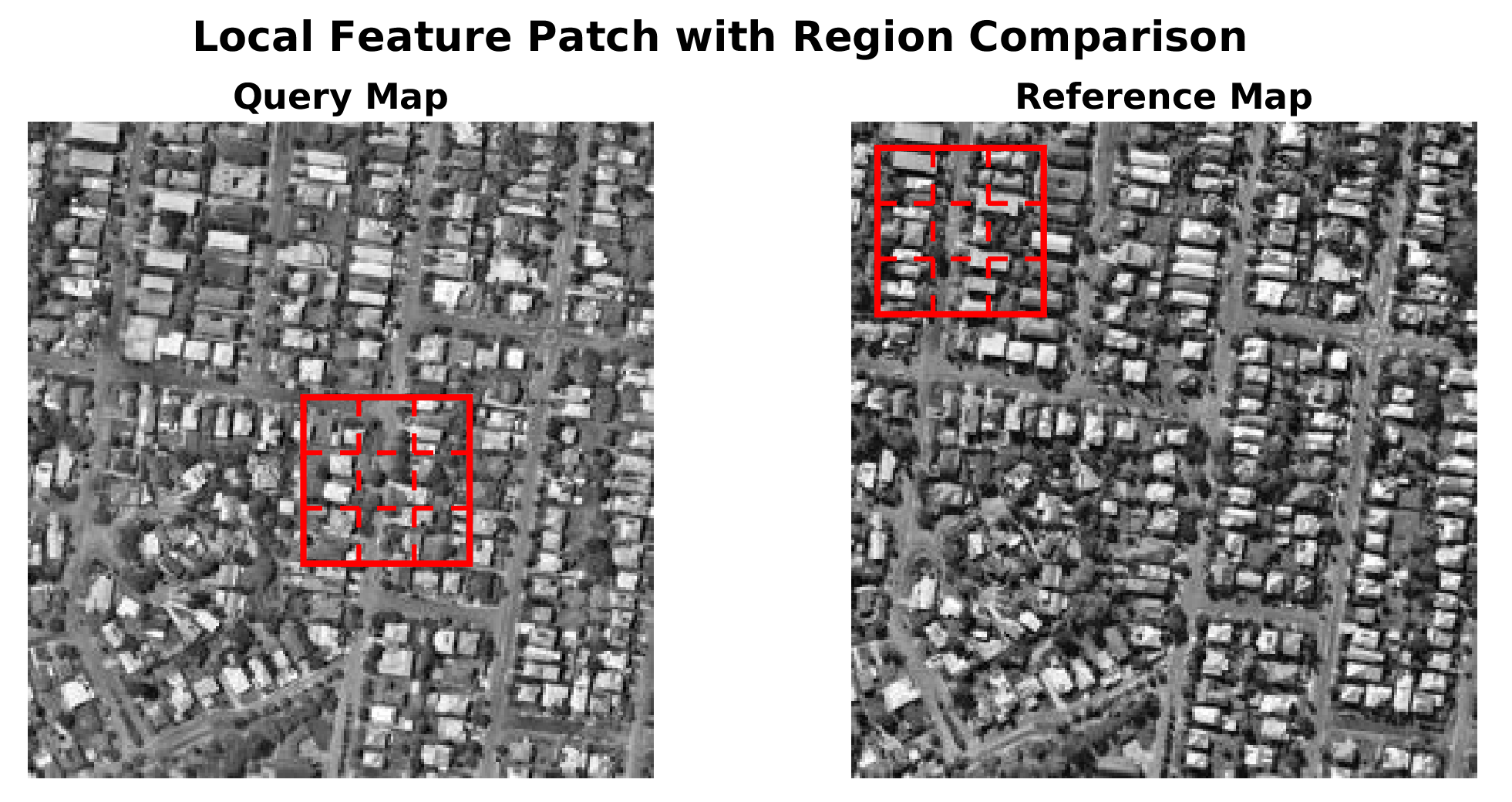}
  \vspace{-0.3cm}
  \caption{An example of the local feature with sub-patch comparison. This technique compares a patch (entire red rectangle) by comparing the corresponding smaller sub-patches. The final metric for a large patch-to-patch comparison is the average percentage of key point inliers across sub-patches. In this work the sub-patch diameter is set to 40 pixels, and we move the patch in increments of 20 pixels. We have used BRISK key points with SURF descriptors, and we only test patch sizes that are integer multiples of the sub-patch size.}
  \label{figure:LocalFeatureTechnique}
\vspace{-0.4cm}
\end{figure}

\textcolor{red}{Within this research our calibration procedure attempts to automatically select the optimal patch radius. We demonstrate the calibration algorithm using two surface-based visual localization techniques, Normalized Cross Correlation (NCC) and local features with sub-patch comparisons. NCC was selected as it has been shown to have relatively good performance within surface-based visual systems, \cite{Kelly:2007, mount2017image, Nourani:2011, aqel2016adaptive}. The local features technique (LFT) is used to demonstrate that the calibration procedure is agnostic to the front-end employed. Figure \ref{figure:LocalFeatureTechnique} shows an example of the local feature with sub-patch comparisons technique. This makes the local feature matching more sensitive to translational shifts and is similar to the regional-MAC descriptor outlined in \cite{tolias2015particular} or the patch verification technique described in \cite{milford2014_Visual}}

\section{Experimental Setup}\label{section:ExperimentalSetup}
This section describes the experimental setup, including the dataset acquisition and key parameter values. All experiments were performed either on a standard desktop running 64-bit Ubuntu 16.04 and MATLAB-2018b or on Queensland's University of Technology's High Performance Cluster running MATLAB-2018b.

\subsection{Image Datasets}
Datasets were either acquired from aerial photography provided by Nearmap, or from road surface imagery collected using a full-frame Sony A7s DSLR. The datasets are summarised in Table \ref{table:Datasets}.

\subsubsection{Aerial Datasets}
\begin{figure*}[!tbhp]
  \centering
  \includegraphics[scale=0.2]{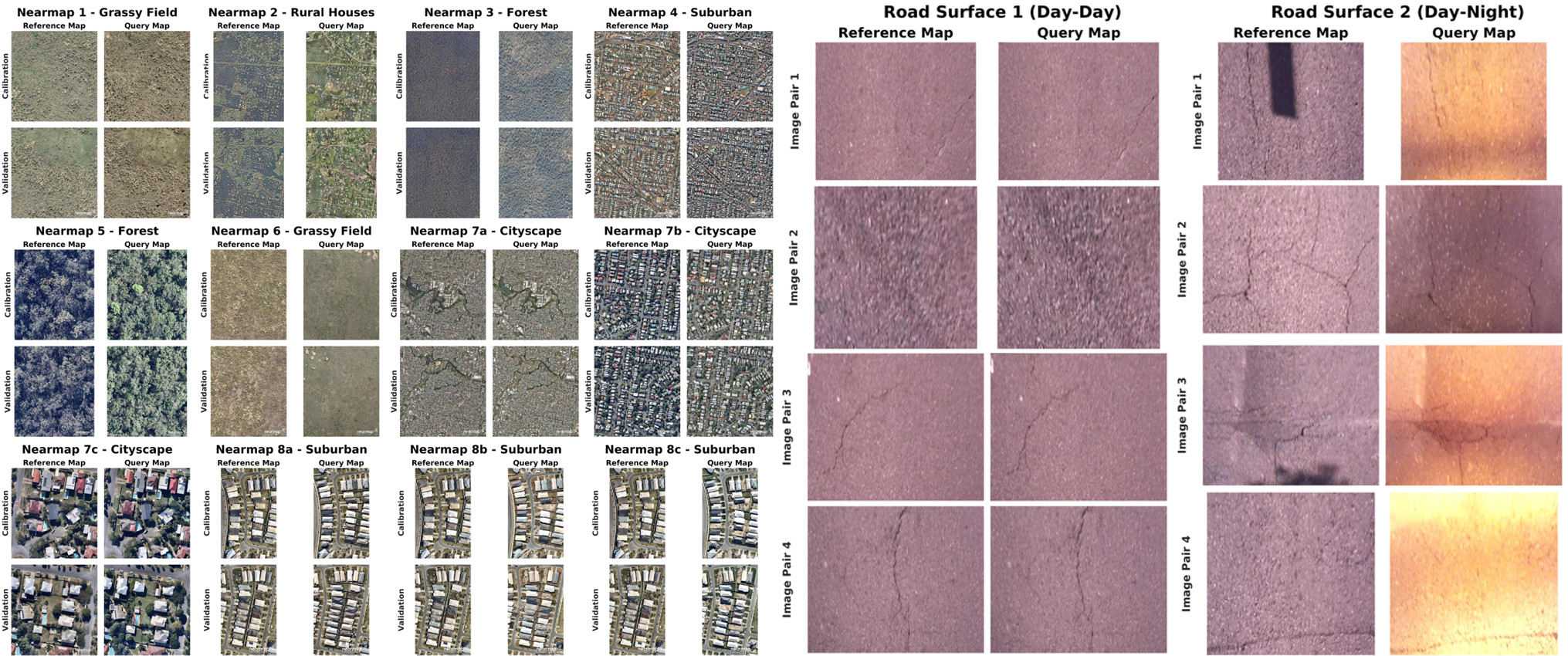}
  \vspace{-0.15cm}
  \caption{The 12 Nearmap reference and query map pairs and 8 image pairs from the Road Surface datasets used in this research. The Nearmap environments vary significantly from grassy fields to urban environments, observed from a range of altitudes and under different appearance changes. The two road surface datasets showing the corresponding reference-query map pairs, with day-day and day-night transitions. The size difference in the images is caused by the manual pixel alignment and cropping procedure.}
  \label{figure:NearmapDatasetExamples}
\vspace{-0.45cm}
\end{figure*}

The aerial datasets were acquired by downloading high-resolution aerial photography provided by Nearmap \cite{nearmap}. 
To ensure suitable dataset variation for validation of our algorithm, the authors collected imagery from forest, field, rural and suburban areas at various altitudes as well as at different qualitative levels of appearance variation. Each Nearmap dataset consists of two pixel aligned images, a reference and a query map. Patches from the query map are compared to the reference map. Figure \ref{figure:NearmapDatasetExamples} shows the reference and query maps for each Nearmap dataset.

The Nearmap Datasets 7a to 7c are from the same location with differing altitudes. \textcolor{red}{Similarly, the Nearmap Datasets 8a to 8c are from the same location with the same reference image, but with different query images with various levels of appearance variation (missing buildings and hue variations)}.

Each Nearmap image was down-sampled to a fixed width while maintaining its aspect ratio. This down-sampling was to increase ease of comparison between different datasets.

\subsubsection{Road Surface Datasets}
The road surface imagery datasets were acquired using a consumer grade Sony A7s, with a standard lens, capturing video while mounted to the bonnet of a Hyundai iLoad van. Three traversals of the same stretch of road were made, two during the day and one at night. Corresponding day-day (Road Surface 1) and day-night (Road Surface 2) frames with significant overlap were then selected, and the corresponding frames manually pixel aligned. This resulted in two datasets, Road Surface 1 and 2. Both datasets have four pixel aligned images, with day-day and day-night images in datasets 1 and 2 respectively. Similarly to the Nearmap datasets, the first image in each image pair is used as the reference map, while the second is used to generate query patches. Figure \ref{figure:NearmapDatasetExamples} shows the four reference and query maps for each Road Surface dataset.

The road surface images were pre-processed, including down-sampling and local patch normalization, to remove the effects of lighting variation and motion blur. This has been shown to improve visual localization performance \cite{milford2012seqslam}.



\begin{table}[!t]
\caption{Datasets}
\label{table:Datasets}
\centering
\begin{tabular}{|c||c||c|}
\hline
\textbf{Dataset Name} & \textbf{Dataset Name} & \textbf{Dataset Name} \\
\hline
Nearmap 1 & Nearmap 2 & Nearmap 3 \\ \hline
Nearmap 4 & Nearmap 5 & Nearmap 6 \\ \hline
Nearmap 7a & Nearmap 7b & Nearmap 7c \\ \hline
Nearmap 8a & Nearmap 8b & Nearmap 8c \\ \hline
Road Surface 1a & Road Surface 1b & Road Surface 1c \\ \hline
Road Surface 2a & Road Surface 2b & Road Surface 2c \\ \hline
\end{tabular}
\vspace{-0.4cm}
\end{table}

\subsection{Parameter Values}
The key parameter values are given in Table \ref{table:KeyParameters}. \textcolor{red}{All parameters were empirically determined over a range of test datasets, and then applied to all experimental datasets. As shown by the results, the system was generally able to select a near optimal patch radius across a range of environment appearances and domains (aerial versus ground-based), even with an almost identical set of parameter values.}
    
\textcolor{red}{The selection of the required Overlapping Coefficient ($O_r$) is a trade off between reducing computational overhead at the risk of reduced localization performance and is dependent on the localization front-end. An initial OVL value can be computed by finding the patch radius that achieves high recall on several test datasets. The remaining parameters, which are mostly dependent on the environment domain and sensor parameters, could also be tuned using exemplary data.}

\begin{table}[!t]
\color{red}
\caption{Key Parameter List for Nearmap and Road Surface Datasets}
\label{table:KeyParameters}
\centering
\begin{tabular}{|c|c|c|c|p{2.1cm}|}
\hline
\textbf{Parameter} & \multicolumn{2}{c|}{\textbf{Nearmap}} & \textbf{Road Surface} & \textbf{Description}\\ \hline
& NCC & LFT & NCC & \\ \hline
$I_{X}$ & 200 & 400 & 100 & Image Width \\ \hline
$N_{X}$ & \multicolumn{2}{c|}{N/A} & 2 & Patch Normalization Radius \\ \hline
$O_{r}$ & 0.005 & 0.0225 & 0.005 &  Required OVL Threshold\\ \hline
$t_{M}$ & \multicolumn{2}{c|}{10} & 5 & True Match Distance Threshold \\ \hline
$N$ & 200 & 100 & 200 & Number of Calibration Samples \\ \hline
$M$ & 1000 & 100 & 1000 & Number of Validation Samples \\ \hline
\end{tabular}
\vspace{-0.4cm}
\end{table}

\section{Experiments and Results}\label{section:Results}

This section presents the results from the various experiments we conducted. To evaluate performance we calculate the recall, as well as a new performance metric which takes into account both recall and computational efficiency. We defined recall as the number of true single frame matches divided by the total number of samples. The second new performance metric is used to test that the calibration procedure does choose the optimal operating point. Optimal performance is defined as maximizing recall with as little computational overhead necessary. This new metric, which we call the max recall to computation efficiency, is given by

\begin{equation}
\color{red}
    M_{i} = 1 - \frac{\sqrt{(P_i - P_{g})^2}}{{arg\, max}_N{(\sqrt{(P_N - P_{g}})^2)}}
\end{equation}

where $M_{i}$ is the max recall to computation efficiency at patch radius $P_i$. $P_g$ and $P_N$ are optimal ground truth patch radius for the dataset and all patch radii used during validation. The ${arg\, max}_N (\sqrt{(P_N - P_{g}})^2$ is used to normalize the distances to be in the range from 0 to 1, while the $1-$ is used to invert the normalized distances so that a higher value means a higher recall to computation efficiency. The optimal ground truth patch radius, $P_g$, is defined as the patch radius which achieves 95\% of the maximum recall for that dataset. This distance metric naturally encodes the recall and computational efficiency into a single value, and it will punish either unnecessary computational overhead or points that achieve poor relative recall. Patch radius is indicative of computational load, as demonstrated in Figure \ref{figure:AveCompTimeAndPatchOverlay}a, which shows that computation time is proportional to patch radius.

\subsection{Automatic Coverage Selection Evaluation}

The first experiment was to investigate the performance of the calibration procedure and test whether it indeed selects the optimal coverage required to maximize localization performance. To evaluate this we ran the calibration routine on a single calibration image that was the same size as and representative of, each Nearmap reference map. We then verified the calibration procedure by testing several patch radii, including the selected patch radius from the calibration routine, on each Nearmap dataset. \textcolor{red}{It should be noted that no image pairs used for calibration are used during validation; and there is no physical overlap between the calibration and validation image pairs in any experiment (see Figure \ref{figure:NearmapDatasetExamples})}. 

\begin{figure}[!tbtp]
  \centering
  \includegraphics[scale=0.20]{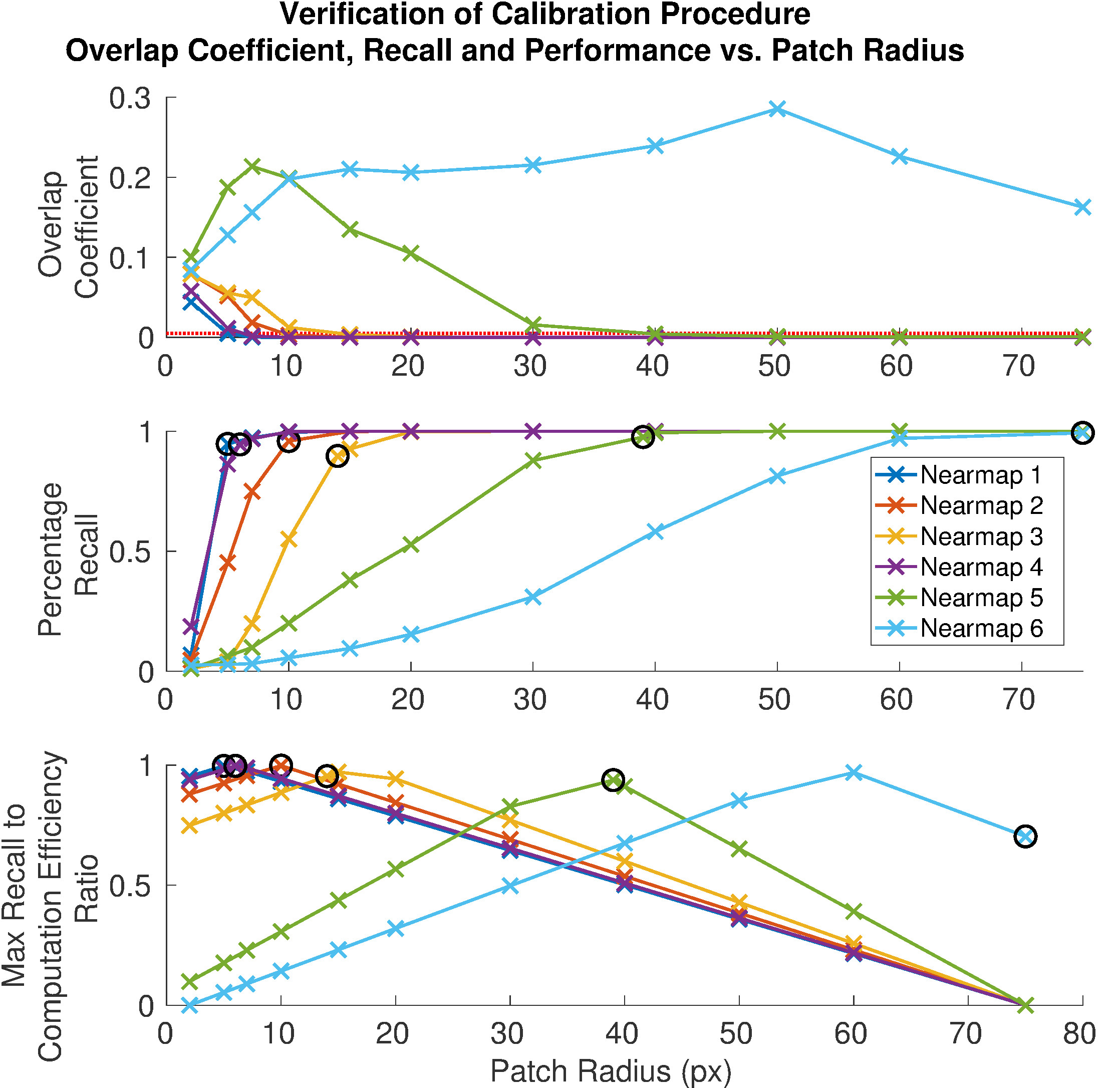}
  \vspace{-0.15cm}
  \caption{Results of the calibration procedure on several Nearmap datasets, optimizing for NCC patch radius. The top plot shows the OVL using Weitzman's measure for the calibration patch radii tested, which is performed on a calibration image. The second and third plot show the percentage recall and max recall to computational efficiency curves for several patch radius, including the selected patch radius, $P_O$, indicated by a black circle, which is performed on the Nearmap dataset images. As can be seen, the calibration procedure consistently selects the patch radius near the top of the max recall to computational efficiency curves, demonstrating its success.}
  \label{figure:ValidationOfCalibrationProcedure}
\vspace{-0.3cm}
\end{figure}

\begin{figure}[!tbtp]
  \centering
  \includegraphics[scale=0.2]{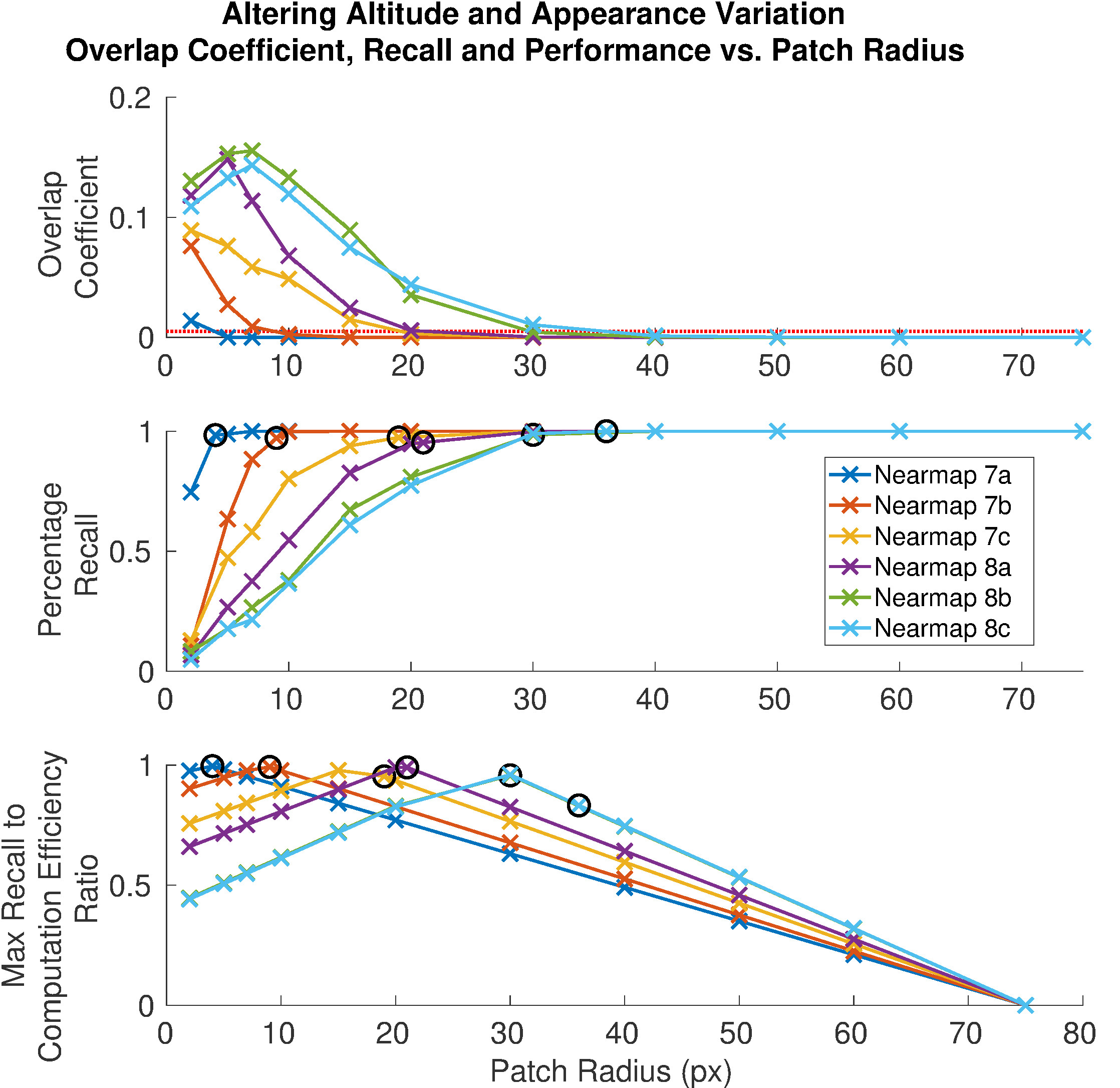}
  \vspace{-0.15cm}
  \caption{Results of the calibration procedure on Nearmap datasets with altitude and appearance variations, datasets 7a-c and 8a-c respectively. As can be seen in the third plot, the calibration consistently picks the near optimal patch size, as indicated by the black circles.}
  \label{figure:AltitudeAndApearanceVariation}
\vspace{-0.25cm}
\end{figure}

To validate the calibration procedure we compute the percentage recall and performance metric for several patch radii on the validation image pairs. The results are shown in Figure \ref{figure:ValidationOfCalibrationProcedure} and \ref{figure:AltitudeAndApearanceVariation}. Figure \ref{figure:ValidationOfCalibrationProcedure} shows the results for Nearmap datasets 1-6. Figure \ref{figure:AltitudeAndApearanceVariation} shows the results for 7a-c and 8a-c which represent various altitudes and appearance variation. 

\textcolor{red}{The Overlap Coefficient for Nearmap 6 does not decay to 0 because the calibration image has an extremely limited amount of unique data (i.e. almost impossible to successfully perform patch localization). Additionally, the validation image does have some unique information which is why 100\% percent recall can be achieved.}

\begin{figure}[!tbtp]
  \centering
  \includegraphics[scale=0.7]{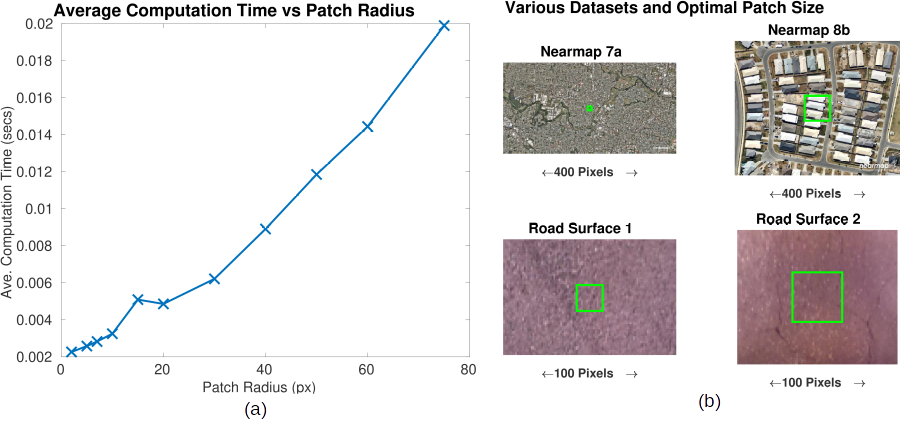}
  \vspace{-0.3cm}
  \caption{(a) Computational profile: the average computation, and hence computational load, is proportional to the patch radius. (b) The optimal visual coverage required is dependent on the data. The rectangles show the optimal patch radius. The optimal patch radius are 4, 30, 7 and 15 pixels for the Nearmap 7a, Nearmap 8b, Road Surface 1 and Road Surface 2 datasets respectively (note that the Nearmap 8b patch radius looks smaller than the Road Surface patch radii because the Nearmap 8b image is 4x larger).}
  \label{figure:AveCompTimeAndPatchOverlay}
\vspace{-0.3cm}
\end{figure}

\begin{figure}[!tbtp]
  \centering
  \includegraphics[scale=0.2]{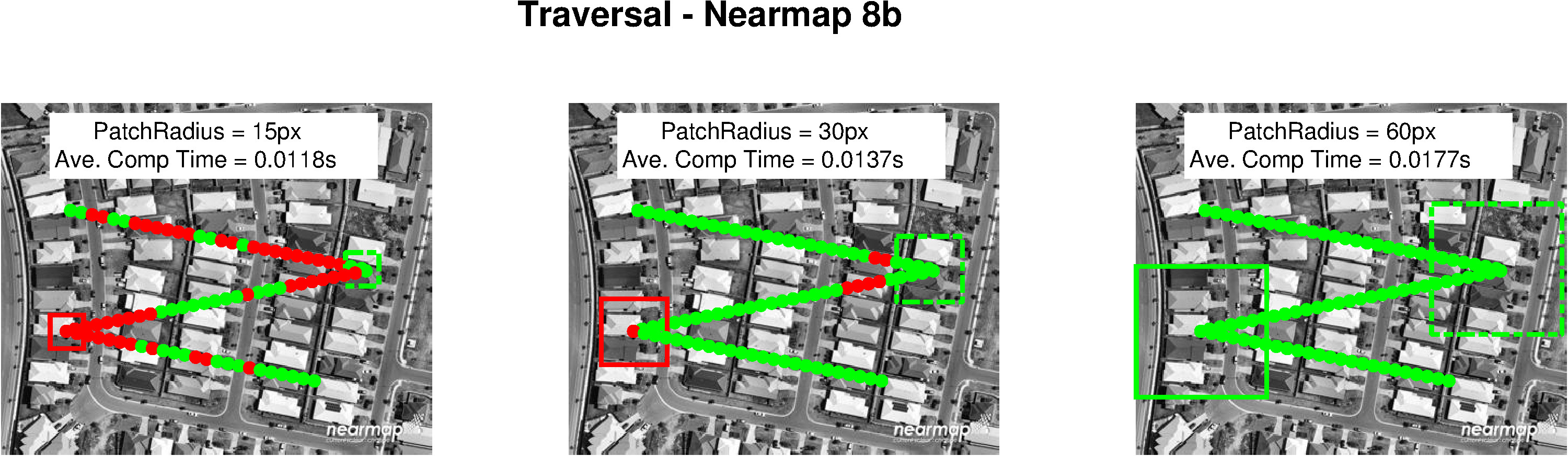}
  \vspace{-0.15cm}
  \caption{A visual indication of the performance of the calibration procedure on a traversal across the Nearmap 8b dataset. As can be seen the optimal patch radius selected by the calibration procedure, 30 pixels, results in almost perfect recall with a much lower computation time per iteration compared to that of the traverse using a 60px patch radius. Each green and red dot indicates the center of successful or unsuccessful localization of a query patch throughout the traverse respectively}
  \label{figure:Nearmap7Traversal}
\vspace{-0.2cm}
\end{figure}

Figure \ref{figure:AveCompTimeAndPatchOverlay}a shows the average computation time is proportional to the patch radius. Additionally, it should be noted that the optimal coverage varies between datasets, as shown in Figure \ref{figure:AveCompTimeAndPatchOverlay}b. In Figure \ref{figure:Nearmap7Traversal} we provide a visual example of a traversal through the Nearmap 8b dataset using the optimal patch radius of 30 pixels, as well as a patch radius above and below. As can be seen, the optimal patch radius results in near perfect recall with minimal computational overhead.

\subsection{Automatic Coverage Selection on a Different Domain}
\begin{figure}[!tbtp]
  \centering
  \includegraphics[scale=0.2]{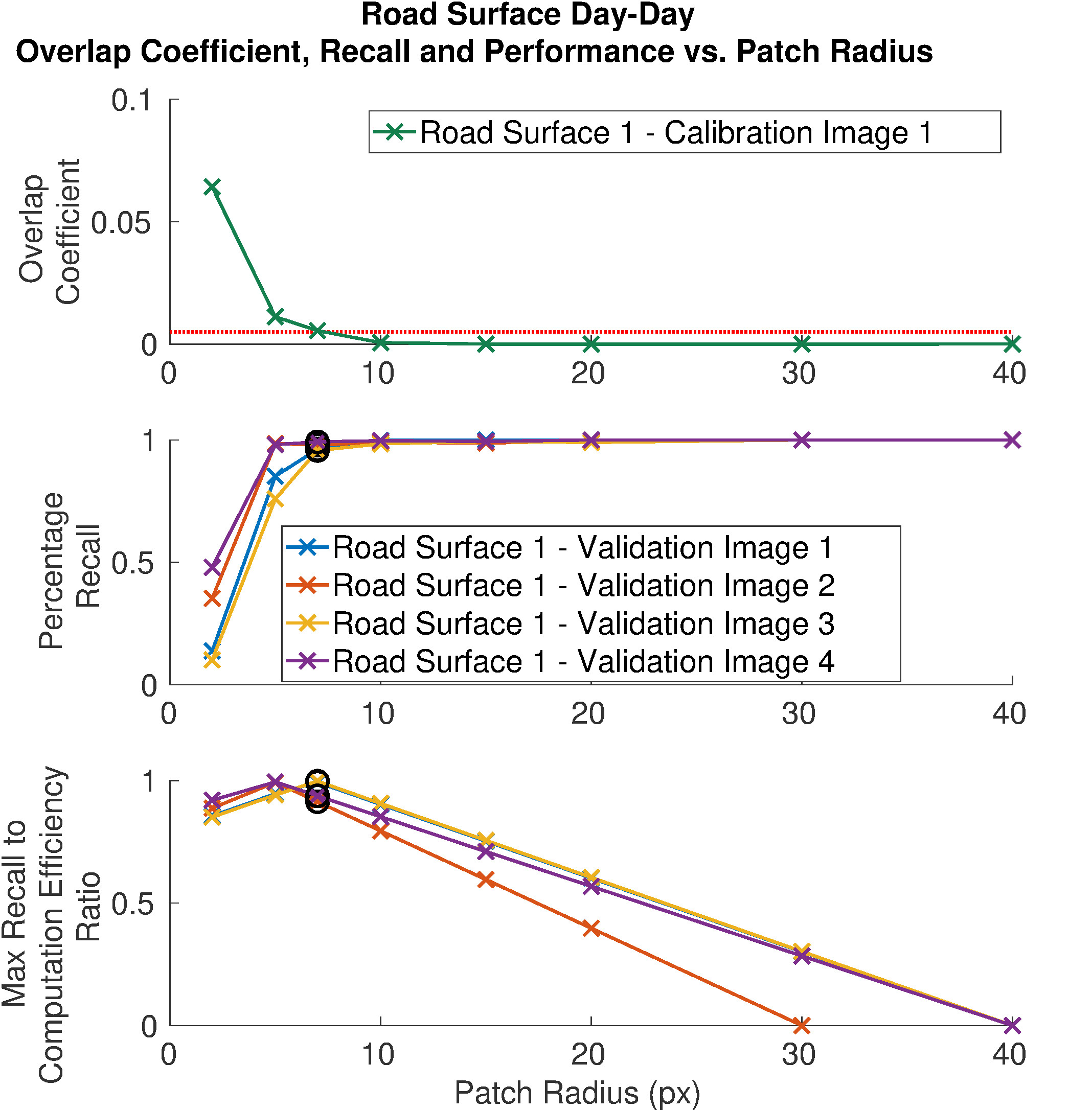}  
  \vspace{-0.15cm}
  \caption{The results of the calibration procedure on the Road Surface 1 dataset (day-day images), which demonstrates that the calibration procedure consistently selects the optimal patch radius within a different data domain.}
  \label{figure:RoadSurfacePlots_DayDay_Single}
\vspace{-0.35cm}
\end{figure}

\begin{figure}[!tbtp]
  \centering
  \includegraphics[scale=0.2]{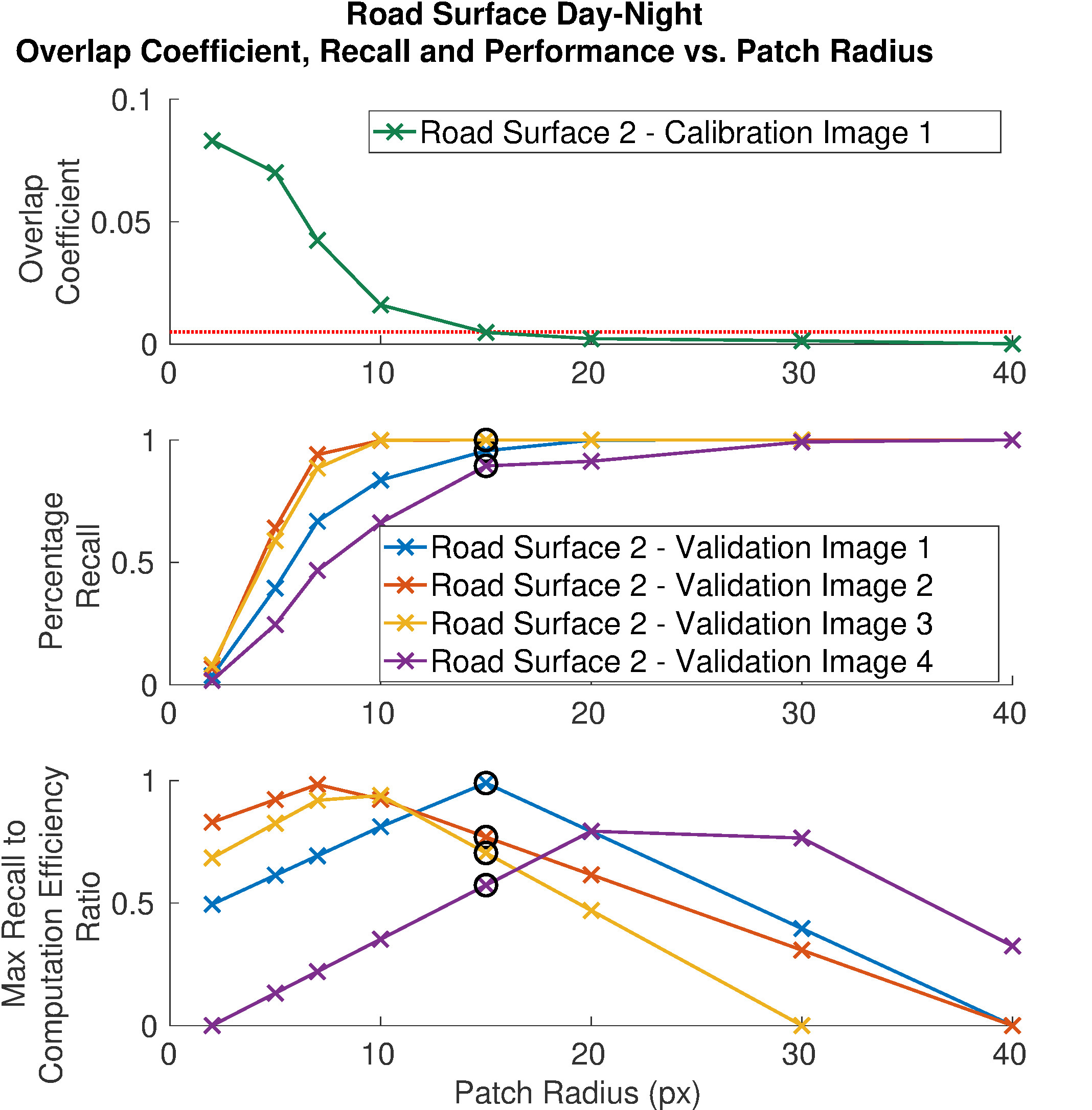} 
  \vspace{-0.15cm}
  \caption{The results of the calibration procedure on the Road Surface 2 dataset (day-night images). The selected patch radius from calibration procedure, which was determined using the first image pair, results in the near optimal performance on the three remaining image pairs within the dataset.}
  \label{figure:RoadSurfacePlots_DayNight_Single}
\vspace{-0.35cm}
\end{figure}

The second experiment investigated how well the automatic selection of the optimal visual coverage worked on a different data domain. \textcolor{red}{For this experiment we used the two road surface datasets. For each dataset, image pair 1 was used for calibration while all four image pairs were used for validation. The results for Road Surface datasets 1 and 2 can be found in Figures \ref{figure:RoadSurfacePlots_DayDay_Single} and \ref{figure:RoadSurfacePlots_DayNight_Single} respectively. Please note we validated on all four images, even though image pair 1 is used for training, to allow us to compare results in the following experiment. We will only discuss the results of image pairs 2 to 4 here.}

As can be seen, the calibration procedure successfully selects the near optimal patch radius in both Road Surface datasets. \textcolor{red}{The slightly lower max recall to computational efficiency performance of the selected patch radius on the Road Surface 2 dataset is due to the fact that the training data in this case was less representative of the deployment data than the other cases. The higher performance on validation image pairs 2 and 3 compared to validation image pair 4 is probably caused by the fact that the unique features in image pairs 2 and 3 (i.e. cracks, identifiable rocks/patterns) are more evenly distributed throughout the entire image. This means that smaller patches have a higher chance of successful localization in validation image pairs 2 and 3, despite any visual variations (i.e. hue) to the calibration image pair.} However, these results still show that the calibration procedure can select an optimal coverage that generalizes to other data (assuming the calibration data is representative of the rest of the dataset). 

\vspace{-0.4cm}

\begin{figure}[!t]
  \centering
  \includegraphics[scale=0.18]{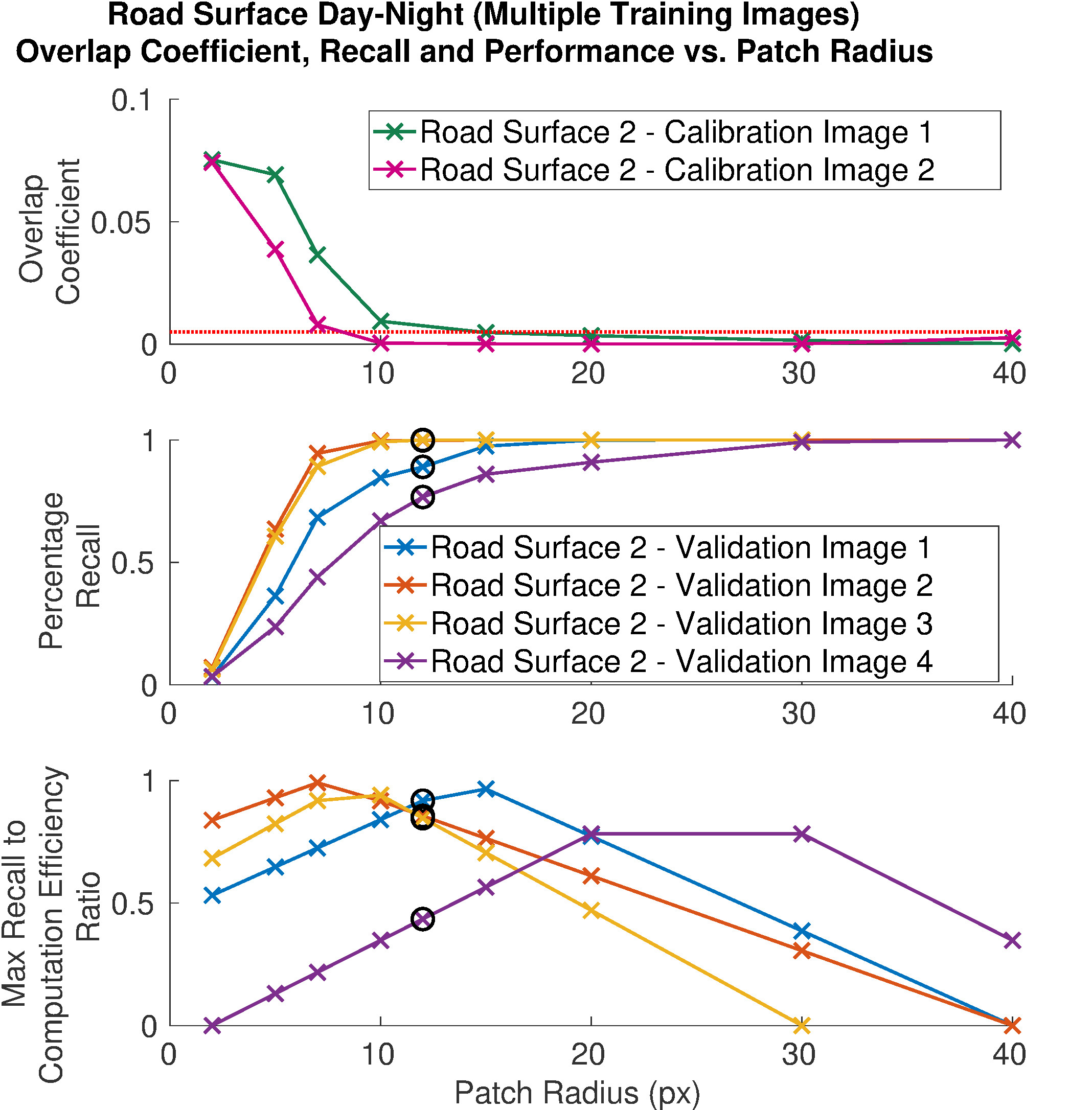}
  \vspace{-0.15cm}
  \caption{The results of the multiple training image experiment performed on Road Surface dataset 2. When comparing to the results from the previous experiment we can see the use of multiple training images improves the overall performance in regards to the max recall to efficiency metric.}
  \label{figure:RoadSurfacePlots_DayNight_Multi}
\vspace{-0.38cm}
\end{figure}

\textcolor{red}{\subsection{Automatic Coverage Selection using Multiple Training Images}}
\textcolor{red}{The previous experiments on Road Surface 2 demonstrate what happens when the training data is not representative of the deployment environment. To mitigate this issue multiple training image pairs can be used. For this experiment we calibrate on image pairs 1 and 2 of the Road Surface 2 dataset and averaged the two optimal patch radii, which were $15$ and $8$ respectively. This average optimal patch radius, $12$, was then validated on all four images. The results are shown in Figure \ref{figure:RoadSurfacePlots_DayNight_Multi}.}

\textcolor{red}{The results show that training on multiple images both positively and negatively affects performance. In the case of images 2 and 3 we can see that the selected patch radius is closer to the peak of the max recall to computational efficiency curve. However, for image pairs 1 and 4 we can see that the selected patch radius has resulted in a decrease on the max recall to computational efficiency curve. For image pairs 1 and 4 this shift on the max recall to computational efficiency curve means the overall recall is decreased (i.e. worse localization performance). In contrast, for image pairs 2 and 3, recall is still maximized but computation efficiency has been increased.  This suggests the averaging of multiple training image pairs does lead to a better overall performance, since there is only a slight decrease in recall performance for image pairs 1 and 4. However, a more sophisticated approach to selecting the optimal patch radius when using multiple image pairs for training may lead to further improvements; this is an avenue for future investigation.}

\begin{figure}[!t]
  \centering
  \includegraphics[scale=0.18]{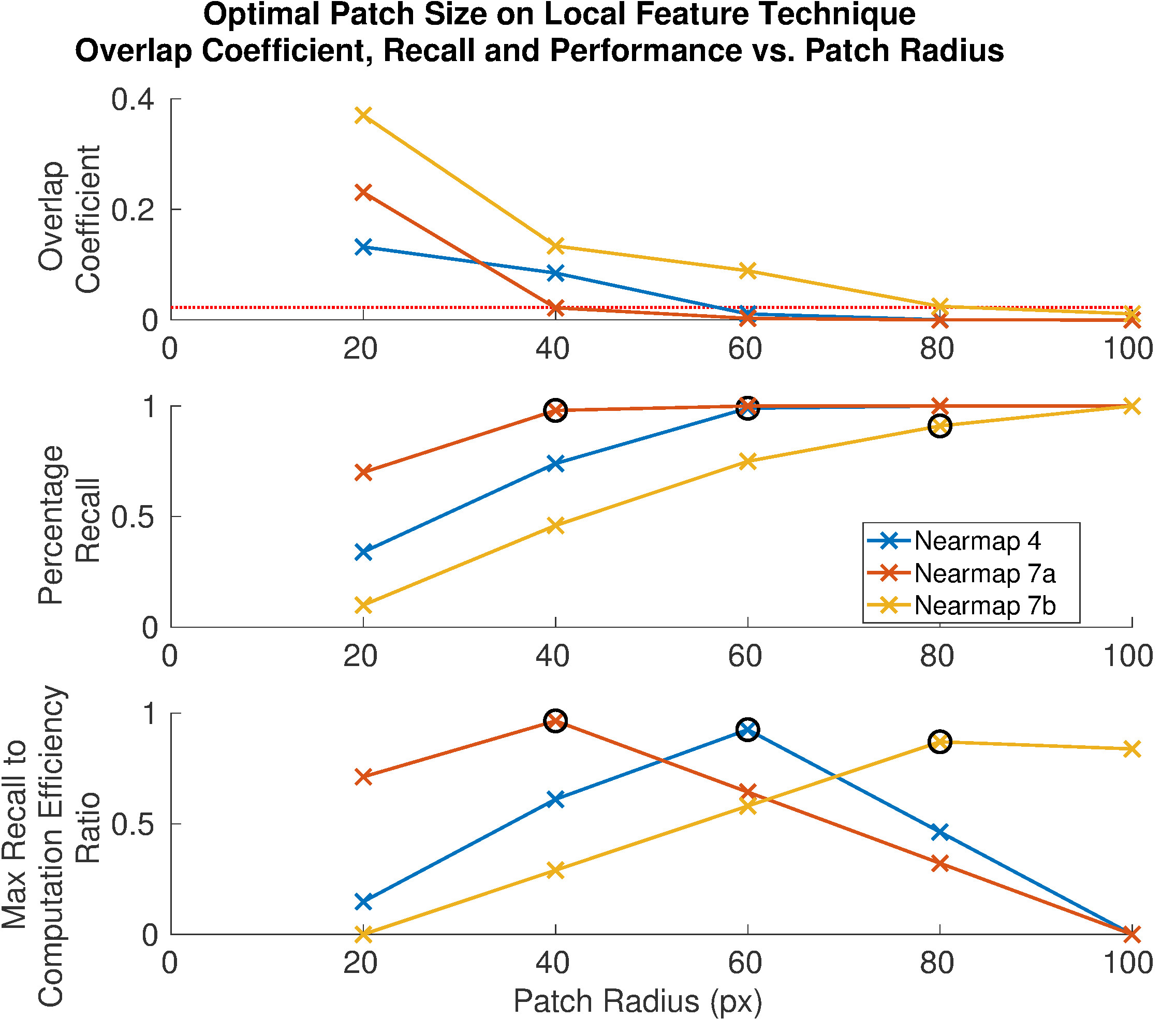}
  \vspace{-0.15cm}
  \caption{The results of using the calibration system with the local feature-based technique. As can be seen the optimal patch radius is correctly selected, showing the proposed system generalizes to other localization front-ends.}
  \label{figure:LocalFeatureTechnique_Results}
\vspace{-0.3cm}
\end{figure}

\textcolor{red}{\subsection{Automatic Coverage Selection Evaluation using a Feature-Based Localization Approach}}
\textcolor{red}{To evaluate the generality of the automatic coverage selection process, we performed a second set of experiments with the local feature-based technique previously described as the localization front-end. Due to the extremely challenging appearance change present in much of the Nearmaps datasets, the feature-based approach only produced competitive performance on datasets 4, 7a and 7b, a result mirroring what has been observed in a range of other feature-based localization systems \cite{milford2014_Visual}. However, for these environments where the underlying front-end was functional, the calibration routine successfully selected the optimal patch radius in all cases, as can be seen in Figure \ref{figure:LocalFeatureTechnique_Results}. These results indicate that the coverage selection process can generalize across different localization front-ends.}

\section{Discussion and Future Work}\label{section:Discussion}

The presented automatic calibration procedure takes a set of aligned imagery from an environment analogous to the deployment domain, and selects the minimum sensor coverage required to achieve optimal localization performance with minimal compute requirements. Experiments run across both aerial and ground-based surface imagery demonstrated that the approach is able to consistently find this optimal coverage amount, even when it varies hugely across application domains and environments.

There are a range of enhancements and extensions that can be pursued in future work. The first is to investigate the potential use of appearance-invariant visual localization algorithms to generate the aligned training data ``on the fly'' at deployment time, removing the need to have training data beforehand \textcolor{red}{and allowing for continuous online calibration}. The second is to investigate other criteria for finding the optimal operating point beyond the implementation used in this research -- such as defining a ``plateau'' threshold in the overlap coefficient curve at which point performance gains diminish with increased sensor coverage.

Thirdly, we have investigated sensor coverage of the environment here but not other properties like sensor resolution. Such properties could likely be optimized through a similar process to the one used here for coverage. \textcolor{red}{Fourthly, the technique has been demonstrated to be agnostic to surface-based visual localization techniques -- it will be interesting to investigate how it performs on other visual localization systems, for example forward-facing cameras.} Additionally, there may be absolute criteria that can be used to determine the optimal coverage for a given environment, again removing the requirement to have training data with aligned imagery. \textcolor{red}{Finally, while the required OVL value is dependent on the localization technique, the heuristically determined OVL thresholds selected appear to be robust across a range of very different datasets and domains, including various image sizes and pre-processing steps. However, a sensitivity analysis would be worth investigating. Additionally, further work into the automatic selection of parameter values as well as a probabilistic interpretation of how to select the OVL value could draw on existing methods, such as \cite{lowry2015building, jacobson2015online}}

Choosing the right camera configuration with respect to mounting and field of view, as well as the operating altitude of an unmanned aerial vehicle, is a critical process both during system design and during deployment operations. We hope that the research presented here will provide an additional tool with which to address these challenges.





\section*{ACKNOWLEDGMENT}

James Mount and Michael Milford are with the Australian Centre for Robotic Vision at the Queensland University of Technology. This work was supported by an ARC Centre of Excellence for Robotic Vision Grant CE140100016, an Australian Research Council Future Fellowship FT140101229 to Michael Milford and an Australia Postgraduate Award and a QUT Excellence Scholarship to James Mount. The authors also appreciate the support and computing resources provided by QUT’s High Performance Centre (HPC).


\bibliographystyle{IEEEtran}
\bibliography{references.bib}

\end{document}